\documentclass[conference,a4paper]{IEEEtran}
\IEEEoverridecommandlockouts
\usepackage{cite}
\usepackage{amsmath,amssymb,amsfonts}
\usepackage{algorithmic}
\usepackage{graphicx}
\usepackage{textcomp}
\usepackage{hyperref}
\usepackage{xcolor}
\usepackage{booktabs}
\usepackage{bm}
\usepackage{tikz}
\usetikzlibrary{positioning}
\usetikzlibrary{shapes.geometric}
\usepackage{algorithm2e}
\usepackage{csquotes}
\usepackage{blindtext}
\usepackage{subcaption}
\usepackage{units}
\newcommand{\Loss}{\mathcal{L}}
\newcommand{\Aug}{\mathcal{T}}
\newcommand{\Data}{\mathcal{D}}
\newcommand{\sT}{\mathcal{T}}
\newcommand{\sQ}{\mathcal{Q}}

\newcommand{\sX}{\mathcal{X}}
\newcommand{\sY}{\mathcal{Y}}
\newcommand{\vc}{\mathbf{c}}
\newcommand{\vu}{\mathbf{u}}
\newcommand{\vv}{\mathbf{v}}
\newcommand{\vq}{\mathbf{q}}
\newcommand{\vp}{\mathbf{p}}
\newcommand{\vone}{\mathbf{1}}
\newcommand{\vx}{\mathbf{x}}
\newcommand{\vy}{\mathbf{y}}
\newcommand{\vz}{\mathbf{z}}
\newcommand{\vtheta}{\boldsymbol{\theta}} 
\newcommand{\vphi}{\boldsymbol{\phi}}

\newcommand{\vbeta}{\boldsymbol{\beta}}

\newcommand{\mC}{\mathbf{C}}
\newcommand{\mQ}{\mathbf{Q}}
\newcommand{\mX}{\mathbf{X}}
\newcommand{\mZ}{\mathbf{Z}}
\def\BibTeX{{\rm B\kern-.05em{\sc i\kern-.025em b}\kern-.08em
    T\kern-.1667em\lower.7ex\hbox{E}\kern-.125emX}}

\begin{document}

\title{TTAPS: Test-Time Adaption by Aligning Prototypes using Self-Supervision}

\author{\IEEEauthorblockN{Alexander Bartler, Florian Bender, Felix Wiewel and Bin Yang}
	\IEEEauthorblockA{\textit{Institute of Signal Processing and System Theory,}
		\textit{University of Stuttgart}\\
		Stuttgart, Germany \\
		Email:    \{alexander.bartler, felix.wiewel, bin.yang\}@iss.uni-stuttgart.de}}


\maketitle
\begin{abstract}
Nowadays, deep neural networks outperform humans in many tasks. However, if the input distribution drifts away from the one used in training, their performance drops significantly. Recently published research has shown that adapting the model parameters to the test sample can mitigate this performance degradation. In this paper, we therefore propose a novel modification of the self-supervised training algorithm SwAV that adds the ability to adapt to single test samples. Using the provided prototypes of SwAV and our derived test-time loss, we align the representation of unseen test samples with the self-supervised learned prototypes. We show the success of our method on the common benchmark dataset CIFAR10-C.
\end{abstract}
\begin{IEEEkeywords}
Test-Time Adaption, Self-Supervision, Test-Time Training, SwAV
\end{IEEEkeywords}

\section{Introduction}
In the last years, deep neural networks are used as the state-of-the-art technology in many applications. However, after deployment, they may be exposed to an input distribution that is different to the one encountered in training. This can be caused by changing weather or lightning conditions for example. Those input distribution shifts often lead to a serious drop in performance \cite{Azulay2019, hendrycks2019benchmarking}. Adversarial attacks, for example, take advantage of the fact that neural networks are sensitive to minimal input distribution shifts which are not even visible for humans. 

To address the problem of a changing input distribution after deployment, many methods focus on special training techniques which lead to models that are robust against adversarial attacks \cite{carlini2017adversarial,Chen_2020_CVPR,Dong_2020_CVPR,Jeddi_2020_CVPR,szegedy2013intriguing, rebuffi2021fixing} or against out-of-distribution samples \cite{shen2021towards, albuquerque2020improving,hendrycks2020many,krueger2020out}.

Another area of research focuses on the adaption of a trained network to a new data domain without supervision which is called Unsupervised Domain Adaption (UDA) \cite{tan2018survey,wang2018deep, wilson2020survey,zhao2020review}. In this case, samples of the shifted input data distribution, the target samples, are available during training but without labels. In combination with the labeled source data samples the model is trained to perform well on the target domain. A special case uses only a single target sample for the adaptation process, which is called one-shot domain adaptation \cite{luo2020adversarial,Benaim2018OneShotUC}.

Both approaches, training a robust model or adapting the model to a new domain, keep the model parameters fixed during testing. Assuming the test sample (e.g. a corrupted image) contains at least some information of the underlying shifted data generating distribution, the adaption of the model parameters using the test sample itself might be enough to at least partially recover the performance of the model. This is called test-time adaption (TTA) \cite{wang2020tent, zhang2021memo} or test-time training (TTT) \cite{sun2020ttt, bartler2021mt3}. 

\subsection{Test-Time Adaption}

\begin{table}[t]
	\centering
	\caption{Comparison of UDA, TTA and TTT (adopted from \cite{wang2020tent})}
	\label{tab:overviewSettings}
	\begin{tabular}{c | c c c c} 
		\toprule
		setting & train loss & test loss \\
		\midrule
		UDA  & $\Loss(\vx_s, y_s) + \Loss(\vx_t)$ & - \\
		
		TTA / TTT \cite{sun2020ttt, bartler2021mt3} & $\Loss(\vx_s, y_s) + \Loss(\vx_s)$ &  $\Loss(\vx_t)$ \\
		fully-TTA \cite{wang2018deep}  & - & $\Loss(\vx_t)$ \\
		\bottomrule
	\end{tabular}
\end{table}

Test-Time Adaption adapts the model, in contrast to e.g. UDA, using the test sample directly. This main difference opens up the possibility to adapt to input distributions coming from different domains or even to non-stationary input data which is only known at test-time. As shown in Table \ref{tab:overviewSettings} this can be realized by directly minimizing a test loss function using the target domain test sample $\vx_t$ without adapting the model nor its training procedure. This is called fully Test-Time Adaption (fully-TTA) in \cite{wang2020tent}. Another possible approach is to extend the model and its training with an auxiliary self-supervised loss function  \cite{sun2020ttt,bartler2021mt3} calculated for the source domain training samples $\vx_s$. During testing only this auxiliary loss function is then minimized on the individual test sample $\vx_t$. In contrast, UDA modifies the training using samples of the target domain which restricts the adaption to a single target domain since it has no flexibility to adapt during testing. One-Shot UDA can be utilized for test-time adaption, but would require the complete training dataset during test-time which would result in a tremendous latency.

Sun et al. \cite{sun2020ttt} showed the effectiveness of a simple self-supervised loss, namely predicting image rotation angles \cite{gidaris2018rot}, to realize TTA. For this, the authors proposed to train the model jointly, supervised and self-supervised. During testing, only the self-supervised loss is minimized using copies of a single test image. Additionally, the authors have shown the effectiveness of this approach in the so called online adaption where the weights are not reset after each test image but adapted continually. 

Bartler et al. \cite{bartler2021mt3} extended the idea of using self-supervision by using meta-learning in their work Meta Test-Time Training (MT3). The authors proposed to meta-train a model such that it learns to adapt in a self-supervised way to unknown test samples. MT3 combines Model-Agnostic Meta-Learning (MAML) \cite{finn2017maml} and the self-supervised training called Bootstrap Your Own Latent (BYOL) \cite{grill2020bootstrap}.

In contrast to this, Wang et al. \cite{wang2020tent} proposed to only adapt during testing without modifying the training procedure or model architecture by minimizing the test entropy (TENT). The authors evaluated the performance of TENT on different datasets for the online setup. Zhang et al. \cite{zhang2021memo} extended and adapted this idea in their work MEMO by minimizing the marginal entropy of augmented copies of a single image such that it is able to adapt to a single sample without any further assumptions (offline TTA). It is also possible to adapt the batch normalization layers of a neural network to work with a single test sample \cite{schneider2020improving}. 



In our work\footnote{A reference implementation of our work is available on GitHub https://github.com/AlexanderBartler/TTAPS}, we will focus on the test scenario where the model is adapted to each test sample individually such that no further assumptions about test-time are required (offline TTA). 

\subsection{Self-Supervision}
Recent successes in self-supervised learning were achieved using contrastive losses \cite{Chen2020Simclr,Chen2020_simclrv2,Oord2018cpcv1,henaf2019cpcv2,caron2020unsupervised}. The key idea of contrastive losses is to maximize the similarity between representations of different augmented versions of the same instance, called positives, while simultaneously minimizing the similarity to other instances, called negatives. 

Besides the success of contrastive losses, previous work on TTA utilized non-contrastive losses \cite{gidaris2018rot,grill2020bootstrap} since during testing only a single instance is available and therefore no negatives are accessible. 

A recent work on self-supervision called \textit{Swapping Assignments between multiple Views of the same image} (SwAV) \cite{caron2020unsupervised} is based on clustering the data while enforcing consistency between different augmentations of the same image by, among other things, matching them to learned prototypes. The learned prototypes can be seen as cluster centers of representations of the data and offer the possibility to be used during testing. 
\\\\
Our contributions are as follows: 
\begin{itemize}
	\item We propose to utilize SwAV to enable Test-Time Adaption by modifying its underlying optimization problem during testing.
	\item We further introduce a simple and effective entropy based regularization to enforce the cluster capability of the learned prototypes. 
	\item We evaluate our method on different benchmark datasets and show that it outperforms many recently proposed related methods on almost all benchmarks.
\end{itemize}

\section{Method}

Our approach leverages the advantages of self-supervision to enable test-time adaptability. Similar to \cite{sun2020ttt, bartler2021mt3} this is realized by jointly minimizing a self-supervised and a supervised loss. During testing, only the self-supervised loss using a single test image is minimized. In the work of Sun et al. \cite{sun2020ttt} simple joint training showed to be effective for TTA. To further enforce the ability to adapt, \cite{bartler2021mt3} proposed to use optimization based meta-learning \cite{finn2017maml} to learn to adapt in a self-supervised way during testing. 

Building on the success of these methods, we propose to jointly minimize the SwAV and supervised loss during training. To enable test-time adaption, we derive a modified test loss by adapting the constraints of the optimization problem used in SwAV for the single instance test-time adaption scenario in order to align the test representation with the best matching prototypes. We call our method \textit{Test-Time Adaption by Aligning Prototypes using Self-Supervision} (TTAPS).

To enhance the ability to cluster the learned prototypes, we further propose to regularize the entropy of the prototypes such that each prototype is clearly related to a single class while on average the prototypes are equally distributed over all classes \cite{grandvalet2005semi,springenberg2015unsupervised}. This can be realized by using a classification head which is used for the minimization of the supervised loss. Learning prototypes using SwAV while jointly minimizing the supervised loss is related to metric based meta-learning \cite{vinyals2016matching,sung2018learning,snell2017prototypical} since we learn prototypes or clusters which are indirectly connected to classes due to the entropy regularization. 
\subsection{Definitions}
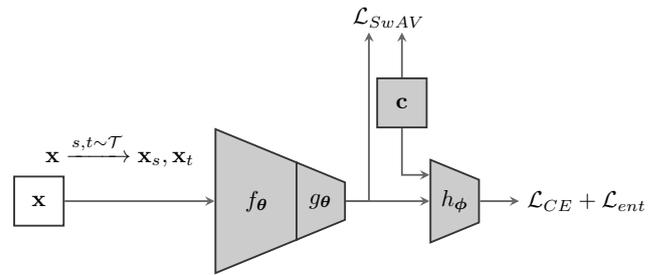
\begin{figure}[!t]
	\centering
	\resizebox{\linewidth}{!}{
		\tikzset{arrow/.style={-stealth, thick, draw=gray!80!black}}
\begin{tikzpicture}

\node[minimum width=0.75cm, minimum height=0.75cm, draw=black!80, thick] (X) at (0.7,0) {$\mathbf x$};

\node[trapezium, draw=black!80, fill=black!20,thick, trapezium angle=67.5, minimum height= 1.25cm, rotate=-90, draw] (F) at (4,0) {\rotatebox{90}{$f_{\vtheta}$}};
\node[trapezium, draw=black!80, fill=black!20,thick, trapezium angle=67.5, minimum height= 0.735cm, rotate=-90, draw] (G) at (4.97,0) {\rotatebox{90}{$g_{\vtheta}$}};

\node[trapezium, draw=black!80, fill=black!20,thick, trapezium angle=67.5, minimum height= 0.735cm, rotate=-90, draw] (Q) at (7,0) {\rotatebox{90}{$h_{\vphi}$}};

\node[minimum width=0.75cm, minimum height=0.75cm, draw=black!80, fill=black!20, thick] (C) at (6.2,1.5) {$\mathbf c$};south

\node[align=left] (Lsup) at (9,0) {$\Loss_{CE}+ \Loss_{ent}$};
\node[align=left] (Lswav) at (6.2,2.7) {};
\node[align=left] (Lswav2) at (6.0,2.8) {$\Loss_{SwAV}$};


%
%
\draw[arrow] (X) -- node[anchor=south, yshift=4mm, xshift=-3mm] {$\vx \xrightarrow{s, t   \sim  \Aug} \vx_s, \vx_t$}(F);
\draw[arrow] (G) -- node[anchor=south] {}(Q);
\draw[arrow] (Q) -- node[anchor=south] {}(Lsup);
\draw[arrow] (C) -- node[anchor=south] {}(Lswav.south);
\draw[arrow] (G.north) -| ([xshift=-5mm]Lswav.south);
\draw[arrow] (C) |-  ([yshift=4mm]Q.south);
\end{tikzpicture}
	}
	\caption{Overview of TTAPS}
	\label{fig:arch}
\end{figure}
Let the training dataset be $\Data^{\text{train}}= \{\vx_n, \vy_n\}_{n=1}^{N}$ with its $N$ inputs $\vx \in \sX$ and the corresponding class label $\vy \in \sY$. During testing, we will consider a single test inputs $\vx_{\text{test}} \in \sX$. 

Following \cite{sun2020ttt, bartler2021mt3}, we modify the architecture as shown in Fig. \ref{fig:arch}. Starting with the convolutional backbone $f_{\vtheta}$, e.g. a ResNet \cite{he2016resnet}, and the following projection head $g_{\vtheta}$, an input sample $\vx_n$ is projected into the $l_2$-normalized lower dimensional projection ${\vz_n \in \lbrace  \vz \in \mathbb{R}^Z \, \vert \, \Vert\vz \Vert_2 = 1 \rbrace}$ where the normalization was suggested by \cite{caron2020unsupervised}. Both, the backbone parameters and projection head parameters are summarized as $\vtheta$. Since our method is based on jointly minimizing the supervised cross-entropy loss $\Loss_{CE}$ and the SwAV loss $\Loss_{SwAV}$, the classification head $h_{\vphi}$ predicts the class label $\hat{\vy}$ based on the projection $\vz_n$ which is needed to minimize $\Loss_{CE}$. The classification head is parameterized by $\vphi$. 

We calculate the prediction directly based on the projection, which is in contrast to \cite{sun2020ttt, bartler2021mt3} where the predictions are calculated based on the output of the backbone and the projections are only used to calculate the self-supervised loss. We further discuss the need to share the projection space between both loss functions in \ref{subsec:entropyloss}.  

One key component of SwAV and our work are the $K$ trainable $l_2$-normalized prototype vectors ${\vc_k \in \lbrace  \vc \in \mathbb{R}^Z \, \vert \, \Vert\vc \Vert_2 = 1 \rbrace}$. A further important component are the augmentation or transformation functions $s,t \sim \sT$ which are randomly sampled from a set of transformations $\sT$ which transforms the input image $\vx_n$ to the augmented views $\vx_{ns},\vx_{nt}$. 

\subsection{SwAV}
In the following we revisit the calculation of the SwAV loss, for further details we refer the reader to \cite{caron2020unsupervised}. Given two different augmentation transformations $s,t \sim \sT$, an input image $\vx_n$ is transformed into $\vx_{ns}$ and $\vx_{nt}$ and projected into $\vz_{ns}, \vz_{nt}$ using $f$ and $g$. The so called codes $\vq_{ns}$ and $\vq_{nt}$ represent the soft-mapping between the projections $\vz_{ns}, \vz_{nt}$ and the $K$ prototypes, further details will follow in \ref{subsec:codes}. In general, the SwAV loss is calculated by the \enquote{swapped} prediction problem \cite{caron2020unsupervised}
\begin{equation}
\Loss = l(\vz_{nt}, \vq_{ns}) + l(\vz_{ns}, \vq_{nt}), 
\end{equation}
where $l(\cdot)$ is the cross-entropy loss between the codes $\vq$ and the softmax of the dot-product of $\vz$ and all prototypes $\vc$
\begin{equation}
\begin{split}
l(\vz_{nt},\vq_{ns}) &= - \sum_k \vq_{ns}^{(k)} \log \vp_{nt}^{(k)}, \\ 
&\text{with} \ \vp_{nt}^{(k)} = \frac{\exp(\frac{1}{\tau}\vz_{nt}^\top \vc_k)}{\sum_{k'} \exp(\frac{1}{\tau}\vz_{nt}^\top \vc_{k'})}
\end{split}
\end{equation}
and $\tau$ is a temperature parameter \cite{wu2018unsupervised}.
Caron et al. motivate the \enquote{swapped} prediction problem by arguing that if $\vz_{ns}$ and $\vz_{nt}$ contain similar information, it should be possible to predict the code of the other representation \cite{caron2020unsupervised}. 

The overall SwAV loss $\Loss_{SwAV}$ of $N$ images is given by
\begin{equation}
\begin{split}
\Loss_{SwAV} = - \frac{1}{N} \sum_{n=1}^{N} \sum_{s,t \sim \sT} \left[ \frac{1}{\tau} \vz_{nt}^\top \mC \vq_{ns} + \frac{1}{\tau} \vz_{ns}^\top \mC \vq_{nt} \right. \\ 
\left.   - \log \sum_{k=1}^{K} \exp\left(\frac{\vz_{nt}^\top \vc_k}{\tau}\right) - \log \sum_{k=1}^{K} \exp\left(\frac{\vz_{ns}^\top \vc_k}{\tau}\right)\right] \label{eq:swav}
\end{split}
\end{equation}
and differentiated w.r.t to the parameters $\vtheta$ and the prototypes ${\mC = \left[\vc_1, \ldots, \vc_K \right]}$.

\subsection{Calculation of the Codes}
\label{subsec:codes}
One key component of the work of \cite{caron2020unsupervised} is the calculation of the codes $\vq$. Given a projection $\vz_i$ and $K$ prototypes $\vc_k$, the $k$-th element $\vq_i^{(k)}$ describes the probability that $\vz_i$ is assigned to the prototype $\vc_k$. 

Given the $K$ prototypes ${\mC = \left[\vc_1, \ldots, \vc_K \right]} \in \mathbb{R}^{Z\times K}$ and a batch of $B$ projections ${\mZ = \left[\vz_1, \ldots,\vz_B \right]} \in \mathbb{R}^{Z \times B}$, the codes ${\mQ = \left[\vq_1, \ldots, \vq_B\right] \in \mathbb{R}^{K\times B}}$ are obtained by solving 
\begin{equation}
 \mQ^* = \underset{\mQ \in \sQ}{ \arg\max} \  \text{Tr}\left(  \mQ^\top \mC^\top \mZ \right) + \epsilon H(\mQ), \label{eq:argmax}
\end{equation}
with the entropy ${H(\mQ) = -\sum_{ij} \mQ_{ij}\log \mQ_{ij}}$. The smoothness of the mapping is controlled by $\epsilon$, normally set to a small value to avoid the trivial solution \cite{caron2020unsupervised}. 

As proposed in \cite{caron2020unsupervised}, the maximization is calculated over the transportation polytope 
\begin{equation}
\sQ = \left\{ \mQ \in \mathbb{R}_{+}^{K \times B} \ \vert \ \mQ \vone_B = \frac{1}{K}\vone_K, \mQ^\top \vone_K = \frac{1}{B}\vone_B\right\}\label{eq:polytope}
\end{equation}
to enforce equal partitioning, such that for each batch, every prototype is selected $\frac{B}{K}$ times on average \cite{caron2020unsupervised}. 

The solution $\mQ^*$ of the constrained optimization problem is given by
\begin{equation}
\mQ^{*} = \text{Diag}(\vu) \exp \left(\frac{\mC^\top \mZ}{\epsilon}\right)\text{Diag}(\vv)
\end{equation}
with the renormalization vectors $\vu \in \mathbb{R}^K, \ \vv \in \mathbb{R}^B$ which have to be iteratively solved using the Sinkhorn-Knopp algorithm \cite{cuturi2013sinkhorn}. Caron et al. suggest to use three iterations \cite{caron2020unsupervised}.  
\subsection{Entropy regularized prototypes}
\label{subsec:entropyloss}
As shown by Sun et al. in \cite{sun2020ttt}, simply minimizing the cross-entropy and a self-supervised loss enables TTA during testing. The authors of MT3 \cite{bartler2021mt3} further proposed to use meta-learning to combine both losses in order to train the model to be adaptable. 

Similar to \cite{bartler2021mt3}, we suggest to connect both loss functions instead of simple joint training to improve the adaption ability during testing where only the self-supervised loss is minimized. 

Previous work already demonstrated the success of entropy regularization techniques   
\cite{grandvalet2005semi,wang2020tent,zhang2021memo} during training as well as during test-time adaption. We modify, in contrast to \cite{sun2020ttt,bartler2021mt3}, our architecture such that the classification prediction is made on the projections $\vz$ instead of the representations $f(\vx)$ (see Fig. \ref{fig:arch}). Due to this shared projection space we are able to regularize the entropy of the prototypes by propagating them through the classification head $h(\cdot)$. Since the prototypes can be interpreted as cluster centers, using an entropy regularization on the predictions leads to an indirect mapping between classes and clusters (prototypes). This assists our TTA since test samples and their projection can be aligned back to prototypes which are indirectly connected to classes by using our modified SwAV test loss presented in \ref{subsec:tta}. 

To regularize the prototypes, we minimize the entropy $H$ of the predictions of prototypes $h(\vc)$ in order to align each prototype uniquely to a specific output of the classification head. 
 Inspired by \cite{zhang2021memo,springenberg2015unsupervised}, we simultaneously maximize the marginal entropy such that all prototypes are linked equally to all classes which avoids collapsing.  
Using Monte-Carlo estimation, our entropy regularization loss $\Loss_{ent}$ is calculated as 
\begin{equation}
\Loss_{ent} = \frac{1}{K} \sum_{k=1}^{K} H(h(\vc_k)) -  H\left(\frac{1}{K} \sum_{k=1}^{K} h(\vc_k)\right) 
\end{equation}
where $H(\cdot)$ is the entropy and $h(\vc_k)$ the prediction of the classification head of $\vc_k$. 
\subsection{TTAPS: Training}
After revisiting the SwAV loss \cite{caron2020unsupervised} and introducing the entropy regularization of the prototypes, we now formulate our training routine.  
Using the cross-entropy loss $\Loss_{CE}(\vy, \hat{\vy})$ and the previously defined SwAV loss and entropy regularization, the training loss of our method TTAPS is given by 
\begin{equation}
\Loss_{TTAPS} = \Loss_{SwAV} + \gamma_1 \Loss_{CE} + \gamma_2 \Loss_{ent}
\end{equation}
where $\gamma_1, \gamma_2 \in \mathbb{R}_+$ are weighing factors. During training, all parts of the loss function are differentiated w.r.t. $\vtheta$. Additionally, $\Loss_{SwAV}$ and $\Loss_{ent}$ are differentiated w.r.t. to the prototypes $\vc$, and $\Loss_{CE}$ and $\Loss_{ent}$ w.r.t. the classification head parameters $\vphi$.

\subsection{TTAPS: Test-Time Adaption}
\label{subsec:tta}
During test-time, we now consider a single test sample $\vx_{test}$. Similar to previous work \cite{sun2020ttt,bartler2021mt3,zhang2021memo} this example is repeated $B_T$ times in order to create a test batch ${\mX_{test} = \left[\vx_{test}, \ldots, \vx_{test}\right]}$. 

Using the copies of the test sample, we minimize only the self-supervised SwAV loss using $P$ gradient steps. For calculating the SwAV loss in \eqref{eq:swav} the codes $\mQ$ are derived  by an optimization over the constraint polytope in \eqref{eq:polytope}, which enforces an equal partitioning between the prototypes and projections. During testing, this constraint does not fit anymore since during testing the batch contains only augmented instances of the same sample and therefore we need to modify the polytope such that all projections could be mapped to a single projection. 

This results in the modified polytope
\begin{equation}
 \hat{\sQ} = \left\{ \mQ \in \mathbb{R}_{+}^{K \times B} \ \vert \ \mQ^\top \vone_K = \vone_B\right\}
\end{equation}
where we now only enforces the columns of $\hat{\mQ}$ to sum up to one. 

To solve \eqref{eq:argmax} using the modified polytope, we follow the original derivation in \cite{cuturi2013sinkhorn}. Due to the modified constraints, the resulting Lagrangian is now simplified to 
\begin{equation}
\mathcal{L}(\hat{\mQ}, \vbeta) = \sum_{kb} -\epsilon \hat{q}_{kb} \log \hat{q}_{kb} + \hat{q}_{kb} \vc_k^\top\vz_b + \vbeta^\top \left(\hat{\mQ}^\top \vone_K - \vone_B\right). \label{eq:lag}
\end{equation}
Solving \eqref{eq:lag} results in the closed form solution
\begin{equation}
\hat{q}_{kb}^* = \frac{\exp \left( \frac{1}{\epsilon}\vc_k^\top \vz_b\right)}{\sum_{k'} \exp \left(\frac{1}{\epsilon}\vc_{k'}^\top \vz_b\right)}
\end{equation}
or 
\begin{equation}
\hat{\vq}_{b}^* = \frac{\exp \left( \frac{1}{\epsilon}\mC^\top \vz_b\right)}{\sum_{k'} \exp \left(\frac{1}{\epsilon}\vc_{k'}^\top \vz_b\right)} \  \text{and} \ \hat{\mQ}^* = \left[\hat{\vq}^*_1, \ldots, \hat{\vq}_B^*\right].
\end{equation}

Using this modified way to calculate the codes, the test SwAV loss $\Loss_{SwAV_{test}}$ can be calculated as in \eqref{eq:swav} and differentiated w.r.t $\vtheta$ (backbone parameters), thus the classification head parameters and prototypes stay fixed. Our TTAPS test-time adaption process for a single sample is shown in Algorithm \ref{alg:tta}. For each test sample, the model is adapted using $P$ gradient steps and afterwards predicts the class label $\hat{\vy} = (h \circ g \circ f)(\vx_{test})$. It is important to mention that in our work the adapted weights are discarded after each test sample and reset to the initially trained parameters (offline TTA). 

\SetKwComment{Comment}{/* }{ */}
\RestyleAlgo{ruled}
\begin{algorithm}[bt]
	\caption{TTAPS: Test-Time Adaption}
	\label{alg:tta}
%
	\textbf{Require:} Pretrained model parameterized by $\vtheta, \vphi$, test sample $x_{test}$, repetition/batch size $B_T$, test steps $P$, test learning rate $\alpha$, transformations $\sT$ \\ 
	
	Initialize parameters $\vtheta^{(0)} = \vtheta$ \\
	\For{$ p = 1,2, \ldots, P$}{
	Repeat $\vx_{test}$ $B_T$ times \\
	Sample tansformations $s,t \sim \sT$ for each copy of $\vx_{test}$ \\
	Apply transformations $s,t$ to each $\vx_{test}$\\
	Adapt model parameters $\vtheta$: \\
	$\vtheta^{(p)} \leftarrow \vtheta^{(p-1)} - \alpha \nabla_{\vtheta} \Loss_{SwAV_{test}} $
	}
	Get final prediction: 
	$\hat{\vy} = (h \circ g \circ f)_{\vtheta^{(P)}}(\vx_{test})$
\end{algorithm}

\begin{figure*}[t!]
	\centering
	\begin{subfigure}[b]{0.45\textwidth}
		\centering
		\includegraphics[width=0.8\linewidth,trim={0em 0em 0em 0em},clip]{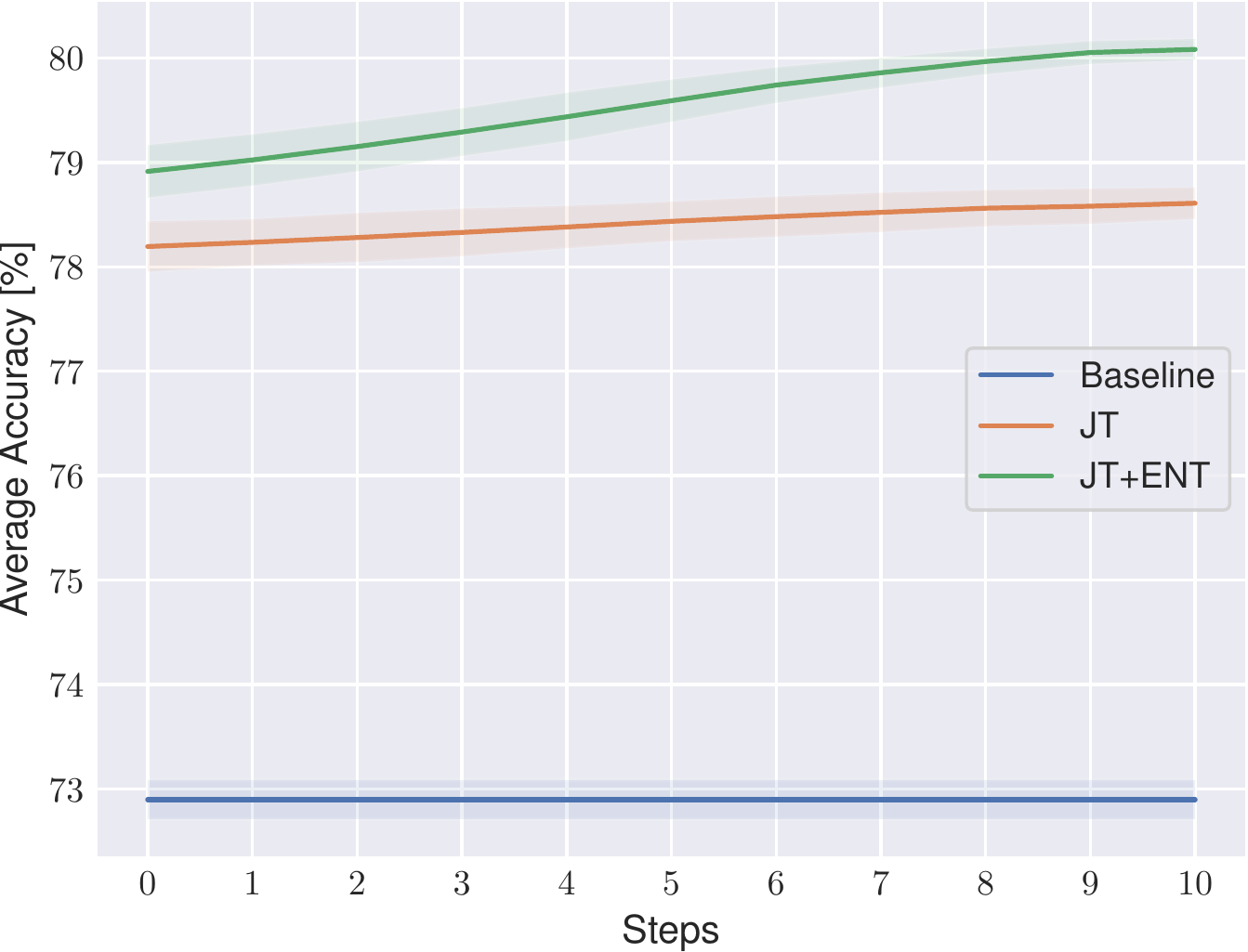}
		\caption{}
		\label{fig:methods1}
	\end{subfigure} 
	\begin{subfigure}[b]{0.465\textwidth}
		\centering
		\includegraphics[width=0.8\linewidth,trim={0em 0em 0em 0em},clip]{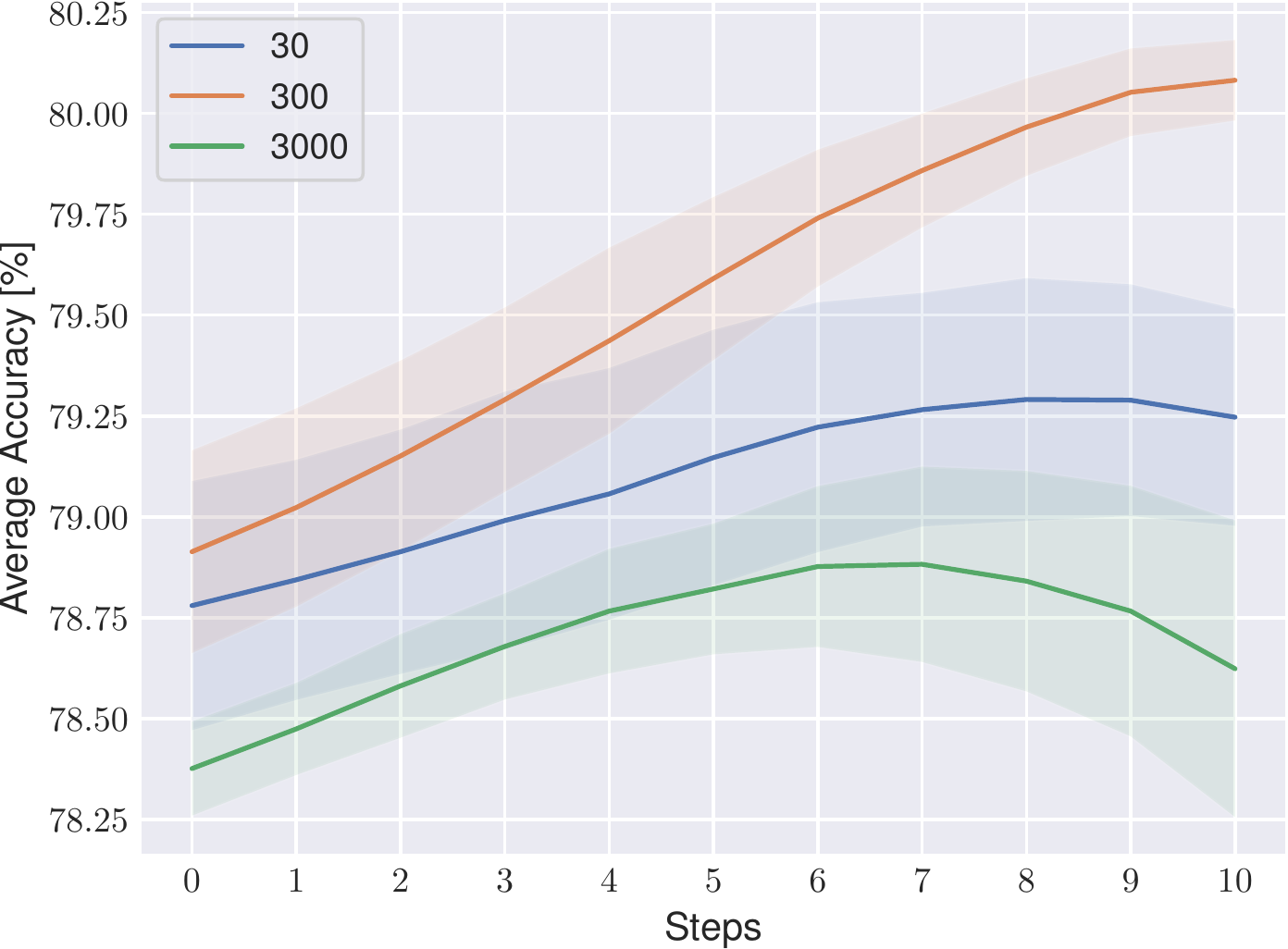}
		\caption{}
		\label{fig:methods2}
	\end{subfigure}
	\caption{Average accuracy over all 15 corruptions with standard deviation using 3 runs. (a): Comparison of entropy regularized joint training (JT+ENT) to purely joint training (JT) and the baseline (supervised only). (b): Results for different number of prototypes of our method TTAPS}
	\label{fig:tta}
\end{figure*}

\section{Experimental Setup}
\subsection{Architecture}
In all our experiments we use a ResNet-26 \cite{he2016resnet} with 32 initial filters as our backbone $f$ in order to have a fair comparison to the previous methods TTT \cite{sun2020ttt}, MT3 \cite{bartler2021mt3} and MEMO \cite{zhang2021memo}. Sun et al. \cite{sun2020ttt} further proposed to use group normalization \cite{wu2018group} instead of batch normalization layers because during testing only a single example is available and therefore no meaningful batch statistics can be estimated. Meanwhile Schneider et al. \cite{schneider2020improving} introduced single-point batch normalization layers, but we will use group normalization layers with 16 groups as well to provide a fair comparison to \cite{sun2020ttt, bartler2021mt3}.  

The projection head $g$ consists of two fully connected layers with 256 and 128 neurons, thus $Z=128$. The supervised head $h$ is a linear layer with an output dimension equal to the number of classes of the dataset.  The number of prototypes is set to $K=300$ in contrast to \cite{caron2020unsupervised} where $K=3000$ was used for the ImageNet dataset \cite{deng2009imagenet}. We will futher discuss the choice of the number of prototypes in Sec. \ref{sec:exp}.

\subsection{Optimization}
As the data augmentation we use similar to the original implementation a random resize crop with random crop size between $0.14$ and $1.0$ of the input dimension and resize it to the original input dimension. We do not consider multi-crop as proposed by \cite{caron2020unsupervised}. We apply color jitter with a strength of $1$ and Gaussian blurring. Further details can be found in the input transformation implementation of \cite{falcon2020framework}. 

During the training of TTAPS  we use a base learning rate of $0.5$ for all our experiments combined with a linear warmup for 10 epochs and cosine decay \cite{loshchilov2016sgdr, misra2020self} as used in the original SwAV training \cite{caron2020unsupervised}. We train for 300 epochs for all our experiments using SGD with a momentum of $0.9$ and weight decay is set to $1 \cdot 10^{-5}$. We use a batch size of $B=256$ and the temperature parameters are set to $\epsilon=0.05$ and $\tau=0.1$ which has also been used in \cite{caron2020unsupervised} for their small batch experiments. The loss weightings are set to $\gamma_1=0.3$ and  $\gamma_2=0.1$. The loss weights are obtained by a small hyperparameter search using the CIFAR10 validation set. 

For TTA we use a test batch size of $B_T=32$ and $P=10$ gradient steps with a learning rate of $0.1$. Furthermore, $\epsilon$ is set to $1.0$ and $\tau=0.75$. Although our method generally adapts all parameters of $f$, preliminary experiments showed only adapting parameters of the last ResNet block is sufficient and slightly more stable.  

\subsection{Datasets}
In our experiments, we mainly show results for the \mbox{CIFAR10} dataset (10 classes) \cite{krizhevsky2009learning}, but to further support our method, we also present results on the more challenging CIFAR100 dataset (100 classes). We split the training set into $40\,000$ images for training and $10\,000$ for validation. 

For testing, and especially the TTA, we use the corrupted CIFAR10/CIFAR100 datasets, CIFAR10-C/CIFAR100-C \cite{hendrycks2019benchmarking} where the CIFAR test images are corrupted with different types of corruptions in different levels of severity. We report all results on the most severe corruption level 5. This dataset is a common benchmark dataset for TTA and was also used in previous works \cite{sun2020ttt,bartler2021mt3,zhang2021memo}. 

\section{Experiments}
\label{sec:exp}
We first analyze the components of TTAPS and show comparisons to our baselines on the CIFAR10-C dataset. Following, we will compare our method against state-of-the-art results on CIFAR10-C and CIFAR100-C.
\subsection{Ablation Study}
\begin{table}[t]
	\centering
	\caption{Loss component overview}
	\label{tab:overviewAblation}
	\begin{tabular}{c | c c c | c} 
		\toprule
		method & $\Loss_{CE}$ & $\Loss_{SwAV}$ & $\Loss_{ent}$ & $\Loss_{SwAV_{test}}$ \\
		\midrule
		Baseline & \checkmark & -  & - & -\\
		\midrule
		JT & \checkmark&  \checkmark& - & -\\
		JT TTA & -& - & -& \checkmark\\
		\midrule
		JT+ENT & \checkmark &  \checkmark&\checkmark &- \\
		\textbf{TTAPS}: JT+ENT TTA & -& - &- & \checkmark\\
		\bottomrule
	\end{tabular}
\end{table}

\begin{figure}[tb!]
	\centering
	\includegraphics[width=0.7\linewidth,trim={1em 1em 1em 1em},clip]{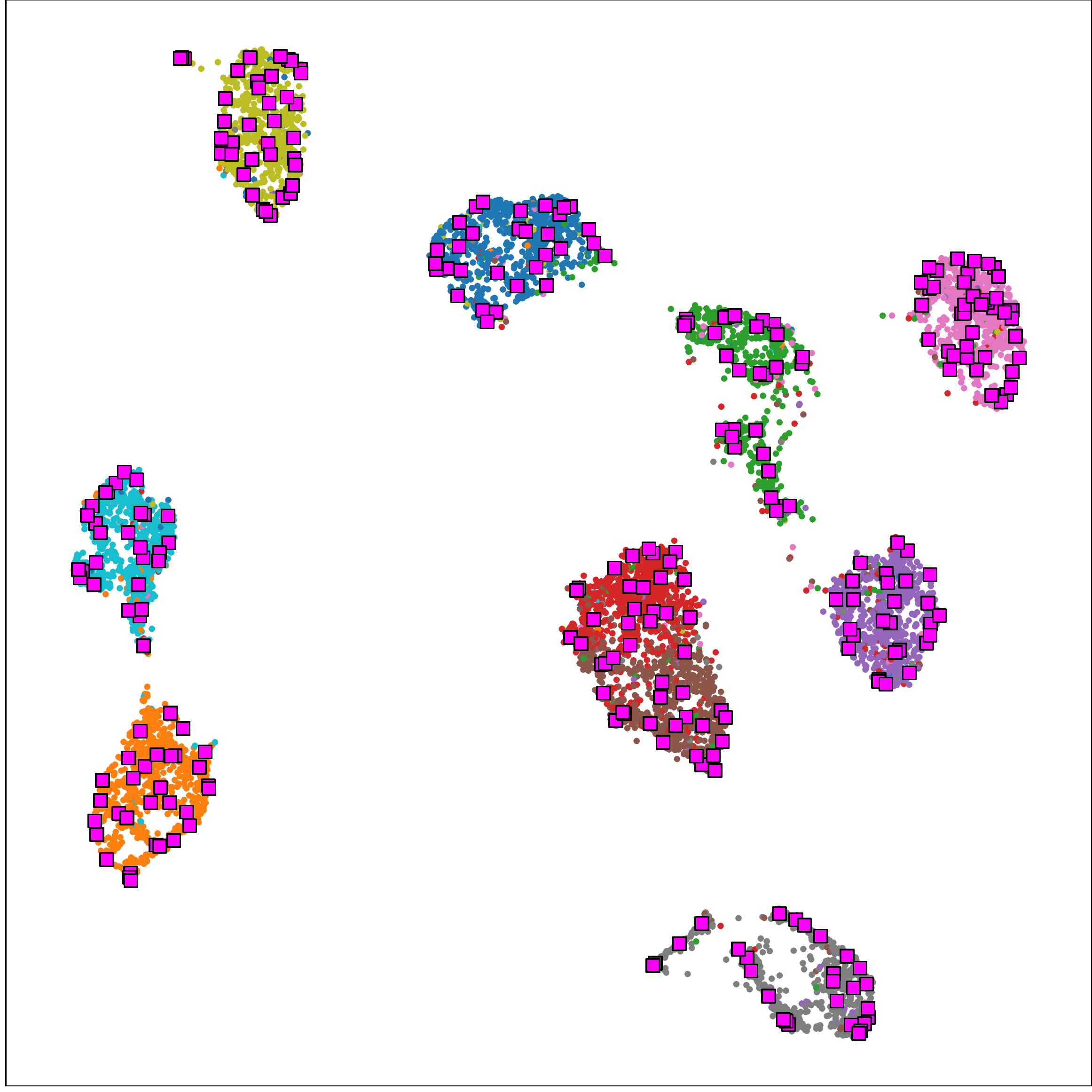}
	\includegraphics[width=0.25\linewidth]{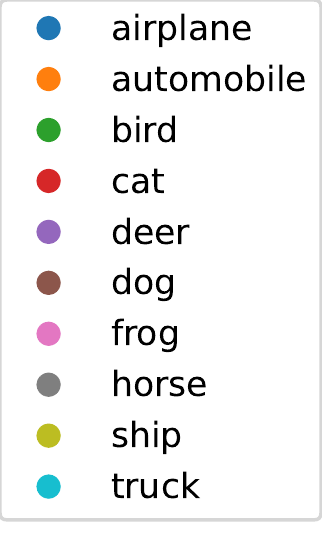}
	
	\caption{Visualization of projections of all 10 classes and prototypes (squares) of CIFAR10 test set using UMAP after training}
	\label{fig:cifar10supervised-swavpem}
\end{figure}

\begin{table*}[th!]
	\centering
	\caption{Comparison of our method to state-of-the art, its baseline, and purely joint training, before and after TTA. Our results (and MT3) on CIFAR10-C are reported with mean and standard deviations of 3 runs}
	\label{tab:ablation}
	\begin{tabular}{c |c c c | c  c c | c c} 
		\toprule
		& TTT\cite{sun2020ttt} & MT3\cite{bartler2021mt3} & MEMO\cite{zhang2021memo} & Baseline & JT & JT TTA  & JT+ENT  & JT+ENT TTA (TTAPS) \\
		\midrule
		brit & $87.8$  & $68.2{\pm} 0.47$  & $88.1$  & $89.8{\pm} 0.30$  & $90.7{\pm} 0.12$  & $91.2{\pm} 0.07$  & $90.9{\pm} 0.08$  & $\mathbf{91.4{\pm}  0.18}$ \\
		contr & $76.1$  & $77.6{\pm} 1.21$  & $71.1$  & $88.4{\pm} 0.15$  & $89.4{\pm} 0.17$  & $89.5{\pm} 0.08$  & $89.4{\pm} 0.14$  & $\mathbf{90.2{\pm}  0.13}$ \\
		defoc & $78.2$  & $84.4{\pm} 0.44$  & $73.6$  & $78.3{\pm} 0.29$  & $88.9{\pm} 0.14$  & $89.0{\pm} 0.15$  & $89.2{\pm} 0.19$  & $\mathbf{89.2{\pm}  0.27}$ \\
		elast & $77.4$  & $76.3{\pm} 1.18$  & $78.9$  & $76.3{\pm} 0.24$  & $79.2{\pm} 0.19$  & $79.6{\pm} 0.36$  & $79.3{\pm} 0.68$  & $\mathbf{79.8{\pm}  0.52}$ \\
		fog & $74.9$  & $75.9{\pm} 1.26$  & $77.2$  & $\mathbf{80.6{\pm}  0.83}$  & $65.4{\pm} 0.93$  & $66.7{\pm} 0.99$  & $67.4{\pm} 1.19$  & $68.2{\pm} 0.88$ \\
		frost & $70.0$  & $81.2{\pm} 0.20$  & $71.7$  & $77.1{\pm} 1.03$  & $83.1{\pm} 0.29$  & $83.5{\pm} 0.13$  & $83.3{\pm} 0.21$  & $\mathbf{84.7{\pm}  0.09}$ \\
		gauss & $54.4$  & $69.9{\pm} 0.34$  & $56.5$  & $55.1{\pm} 0.35$  & $66.4{\pm} 1.07$  & $66.6{\pm} 1.02$  & $68.1{\pm} 0.70$  & $\mathbf{71.0{\pm}  0.50}$ \\
		glass & $53.9$  & $66.3{\pm} 1.24$  & $55.6$  & $57.2{\pm} 0.14$  & $68.4{\pm} 0.46$  & $67.5{\pm} 0.38$  & $\mathbf{68.7{\pm}  1.34}$  & $68.6{\pm} 0.92$ \\
		impul & $50.0$  & $58.2{\pm} 1.25$  & $56.7$  & $52.8{\pm} 0.59$  & $54.7{\pm} 0.71$  & $57.8{\pm} 0.18$  & $56.4{\pm} 1.19$  & $\mathbf{62.4{\pm}  0.55}$ \\
		jpeg & $72.8$  & $77.3{\pm} 0.26$  & $78.3$  & $73.2{\pm} 0.72$  & $81.9{\pm} 0.43$  & $82.4{\pm} 0.33$  & $82.5{\pm} 0.12$  & $\mathbf{83.6{\pm}  0.33}$ \\
		motn & $77.0$  & $77.2{\pm} 2.37$  & $74.9$  & $\mathbf{83.3{\pm}  0.51}$  & $81.3{\pm} 0.29$  & $81.6{\pm} 0.20$  & $82.3{\pm} 0.35$  & $82.0{\pm} 0.51$ \\
		pixel & $52.8$  & $72.4{\pm} 2.29$  & $57.2$  & $62.7{\pm} 3.39$  & $82.3{\pm} 0.37$  & $81.2{\pm} 0.76$  & $\mathbf{82.5{\pm}  0.29}$  & $81.8{\pm} 0.36$ \\
		shot & $58.2$  & $70.5{\pm} 0.72$  & $60.2$  & $58.4{\pm} 0.91$  & $69.2{\pm} 0.95$  & $69.8{\pm} 1.09$  & $70.5{\pm} 0.18$  & $\mathbf{73.9{\pm}  0.15}$ \\
		snow & $76.1$  & $79.8{\pm} 0.63$  & $79.1$  & $81.2{\pm} 0.24$  & $83.3{\pm} 0.17$  & $84.0{\pm} 0.16$  & $83.7{\pm} 0.75$  & $\mathbf{85.1{\pm}  0.45}$ \\
		zoom & $76.1$  & $81.3{\pm} 0.58$  & $75.0$  & $79.2{\pm} 0.34$  & $88.8{\pm} 0.14$  & $88.9{\pm} 0.22$  & $\mathbf{89.4{\pm}  0.04}$  & $89.3{\pm} 0.09$ \\
		\midrule
		Avg. & $69.0$  & $75.6{\pm} 0.30$  & $70.3$  & $72.9{\pm} 0.19$  & $78.2{\pm} 0.24$  & $78.6{\pm} 0.15$  & $78.9{\pm} 0.25$  & $\mathbf{80.1{\pm}  0.10}$ \\
		\bottomrule
		
	\end{tabular}
\end{table*}

A component overview of our method and its baseline is shown in Table \ref{tab:overviewAblation}. The baseline training is using only the cross-entropy loss $\Loss_{CE}$ without any self-supervision or regularization. For a fair comparison, we use the same architecture including the projection head $g$ and training techniques e.g. the same warmup and learning rate decay. We set the base learning rate to $0.1$. For purely joint training (JT), we use exactly the same parameter as for TTAPS, but without the entropy regularization loss $\Loss_{ent}$. If test-time adaption is applied to JT, we call it JT+TTA, where the modified test SwAV loss $\Loss_{SwAV_{test}}$ is used for adapting the model to the corrupted test samples. Finally, we show the results for our method TTAPS (JT+ENT) before and after test-time adaption (JT+ENT TTA).

After the training of our method (JT+ENT), the projections $\vz_n$ of the CIFAR10 test set are visualized with the learned prototypes $\vc_k$ in Fig. \ref{fig:cifar10supervised-swavpem} using UMAP \cite{mcinnes2018umap}. It can be seen that the prototypes are aligned with different parts of each cluster which highlights the effectiveness of the shared projection space in combination with our loss functions. During testing of a corrupted sample, which will maybe not projected directly into a cluster, our test loss $\Loss_{SwAV_{test}}$ aligns it to the prototypes by adapting the model parameters. This leads to an improved accuracy since the prototypes are initially learned to be aligned with classes due to the entropy regularization.

In Fig. \ref{fig:tta}(a), we show the test-time adaption behavior of TTAPS against the simple baseline and purely JT+TTA. Therefore, the averaged accuracy on all 15 corruptions over the number of test-time gradient steps is shown with mean and standard deviation of 3 runs. Both, JT and JT+ENT, have a clearly better initial robustness against the corruptions which is due to the applied self-supervised loss and its stronger data augmentation. A similar behavior was reported in \cite{bartler2021mt3} where BYOL \cite{grill2020bootstrap} was used as self-supervision. The ability to adapt is given with and without the entropy regularization as the averaged accuracy is monotonically increasing over the number of gradient steps in both cases. As we motivated in \ref{subsec:entropyloss}, the entropy regularization of the prototypes boosts the ability to adapt by a large margin. This highlights that the entropy regularization leads to better prototypes if they are used for aligning unknown test samples during test-time adaptation.       

The prototypes are a key component of TTAPS. Therefore, we analyze the impact on the number of prototypes on the test-time adaption of TTAPS in Fig. \ref{fig:tta}(b). We compare the results for 30, 300 and 3000 prototypes using the average accuracy with mean and standard deviation of 3 runs. Caron et al. \cite{caron2020unsupervised} mentioned that SwAV is not sensitive to the number of prototypes as long as the number is high enough. TTAPS, in contrast, seems to be sensitive to the number of prototypes. If 30 or 3000 prototypes are used, still the average accuracy is increasing for a few steps but also decrease for a higher number of gradient steps where as with 300 prototypes the average accuracy is monotonically increasing with an even lower standard deviation. 

The detailed results with mean and standard deviation of 3 runs for each corruption of our method and its baselines are shown in Table  \ref{tab:ablation}. With a mean accuracy of $\unit[72.9]{\%}$ we provide a strong baseline in means of robustness to corruptions. Adding the SwAV loss during training increases the average accuracy to $\unit[78.2]{\%}$. Applying TTA to purely joint training increases the accuracy slightly by $\unit[0.4]{\%}$. In contrast to this, using in addition the entropy regularization leads to $\unit[78.9]{\%}$  and an additional increase of $\unit[1.2]{\%}$ after TTA. For only 4 of 15 corruption TTA (TTAPS) leads to a small drop in performance. In contrast to this, for an input corrupted by e.g. impulse noise (impul) our method increases the accuracy from $\unit[56.4]{\%}$ to $\unit[62.4]{\%}$ which is relative improvement of $\unit[10.6]{\%}$. Again, this is achieved by adapting to each single test sample individually. 

\subsection{Comparison to state-of-the-art}
\begin{table*}[t!]
	\centering
	\caption{Comparison to state-of-the art on CIFAR100-C. Results are presented with mean and standard deviations of 3 runs}
	\label{tab:ablation3}
	\begin{tabular}{c | c |c c c c } 
		\toprule
		& MT3 \cite{bartler2021mt3}  & JT & JT TTA & JT+ENT & JT+ENT TTA (TTAPS) \\
		\midrule
		brit & $52.2{\pm} 0.44$  & $61.1{\pm} 0.15$  & $61.1{\pm} 0.05$  & $62.3{\pm} 0.06$  & $\mathbf{62.7{\pm}  0.30}$ \\
		contr & $31.6{\pm} 1.53$  & $54.7{\pm} 0.29$  & $54.3{\pm} 0.40$  & $\mathbf{56.3{\pm}  0.41}$  & $56.3{\pm} 0.57$ \\
		defoc & $55.0{\pm} 0.55$  & $60.7{\pm} 0.31$  & $60.7{\pm} 0.24$  & $61.2{\pm} 0.74$  & $\mathbf{61.6{\pm}  0.50}$ \\
		elast & $44.2{\pm} 0.81$  & $45.5{\pm} 0.31$  & $45.7{\pm} 0.43$  & $46.3{\pm} 0.79$  & $\mathbf{46.8{\pm}  0.80}$ \\
		fog & $33.3{\pm} 0.45$  & $30.3{\pm} 0.30$  & $30.5{\pm} 0.35$  & $\mathbf{34.2{\pm}  0.69}$  & $34.2{\pm} 0.70$ \\
		frost & $45.5{\pm} 1.00$  & $46.4{\pm} 0.80$  & $47.0{\pm} 0.80$  & $48.1{\pm} 1.16$  & $\mathbf{48.5{\pm}  1.20}$ \\
		gauss & $\mathbf{32.8{\pm}  0.84}$  & $25.4{\pm} 0.69$  & $25.2{\pm} 0.57$  & $27.1{\pm} 0.57$  & $25.8{\pm} 0.67$ \\
		glass & $\mathbf{33.0{\pm}  0.93}$  & $32.8{\pm} 0.53$  & $32.8{\pm} 0.42$  & $32.8{\pm} 1.43$  & $32.2{\pm} 1.33$ \\
		impul & $\mathbf{18.4{\pm}  0.09}$  & $16.1{\pm} 0.38$  & $16.2{\pm} 0.30$  & $17.8{\pm} 0.96$  & $18.4{\pm} 1.21$ \\
		jpeg & $42.7{\pm} 0.46$  & $46.6{\pm} 0.55$  & $47.6{\pm} 0.54$  & $48.5{\pm} 0.57$  & $\mathbf{50.2{\pm}  0.66}$ \\
		motn & $45.4{\pm} 0.81$  & $52.4{\pm} 0.20$  & $52.3{\pm} 0.15$  & $53.4{\pm} 0.48$  & $\mathbf{53.4{\pm}  0.55}$ \\
		pixel & $41.2{\pm} 2.06$  & $49.9{\pm} 0.61$  & $49.3{\pm} 0.88$  & $\mathbf{52.3{\pm}  1.03}$  & $52.0{\pm} 0.88$ \\
		shot & $\mathbf{33.1{\pm}  1.41}$  & $28.0{\pm} 1.06$  & $27.7{\pm} 0.97$  & $29.3{\pm} 0.66$  & $28.8{\pm} 1.06$ \\
		snow & $43.7{\pm} 1.12$  & $48.5{\pm} 0.53$  & $49.3{\pm} 0.53$  & $49.7{\pm} 0.74$  & $\mathbf{50.5{\pm}  0.74}$ \\
		zoom & $52.0{\pm} 0.62$  & $61.7{\pm} 0.39$  & $61.9{\pm} 0.44$  & $61.3{\pm} 0.71$  & $\mathbf{62.0{\pm}  0.68}$ \\
		\midrule
		Avg. & $40.3{\pm} 0.27$  & $44.0{\pm} 0.18$  & $44.1{\pm} 0.17$  & $45.4{\pm} 0.39$  & $\mathbf{45.6{\pm}  0.46}$ \\
		\bottomrule
		
	\end{tabular}
\end{table*}

We compare our results to previous state-of-the art methods, namely TTT \cite{sun2020ttt}, MT3 \cite{bartler2021mt3} and MEMO \cite{zhang2021memo}. All results are reported for the same architecture ResNet-26. TTT utilizes similar to our work a self-supervised loss, but a rather simple rotation prediction \cite{gidaris2018rot}. In contrast to this, MT3 utilizes the strong self-supervised BYOL \cite{grill2020bootstrap} loss in combination with meta-learning. Both methods, TTT and MT3, are comparable to our method by means of adapting the training using a self-supervised loss as shown in Table \ref{tab:overviewSettings}. To further compare our method, we show the results of MEMO which is a fully-TTA method. This means that the model is adapted during testing without modifying its training routine. Therefore, the comparison of TTT, MT3 and our TTAPS to MEMO is to be taken with caution. We show the results after test-time adaption and are thus comparable with TTAPS (JT+ENT TTA).

The comparison is shown in Table \ref{tab:ablation}. TTT and MEMO have a similar overall performance with $\unit[69]{\%}$ and $\unit[70.3]{\%}$. MT3, in contrast, leads to an average accuracy over all corruptions of $\unit[75.6]{\%}$. Our method TTAPS outperforms all previous methods by a large margin. If we compare our method to MT3 where a similarly strong self-supervised loss was used, we observe that our method utilizes its self-supervised loss more effectively as our JT+ENT (before adaption) which leads already to a higher average accuracy. Additionally, applying TTA using our test loss leads to a large improvement. Furthermore, our method is computationally more efficient than MT3 since we do not need second order derivatives during training like MT3. 

To further evaluate TTAPS, we show the results on the more challenging dataset CIFAR100-C in Table \ref{tab:ablation3} where we compare only to MT3 since TTT and MEMO do not evaluate their method on CIFAR100-C. The hyperparameters for our method are exactly the same as for CIFAR10. We show our results with and without entropy regularization (ENT) before and after TTA. Despite the improvement of TTA being not that large for this challenging dataset, TTAPS still outperforms MT3 again by a large margin.

\section{Conclusion}
In this paper, we propose a novel modification of the self-supervised SwAV loss which enables the ability to adapt during test-time using a single test sample. This is realized by adapting the constraints of the SwAV optimization problem. Furthermore, we propose to regularize the entropy of the learned prototypes. We discuss and analyze the components of our method TTAPS and compare it to state-of-the-art results in single sample test-time adaption. Using the same backbone architecture, our method improves the results by a large margin on the CIFAR10-C and CIFAR100-C dataset.

\bibliographystyle{IEEEtran}
\bibliography{refs}

\end{document}